\documentclass[final]{cvpr}

\usepackage{times}
\usepackage{epsfig}
\usepackage{graphicx}
\usepackage{amsmath}
\usepackage{amssymb}
\usepackage{multirow}
\usepackage{comment}

\usepackage[pagebackref=true,breaklinks=true,colorlinks,bookmarks=false]{hyperref}

\def\ourmodel{GCoNet}

\def\cvprPaperID{1094} 
\def\confYear{CVPR 2021}
\graphicspath{{./figure/}}
\DeclareGraphicsExtensions{.jpg,.pdf,.png}
\begin{document}
\bibliographystyle{unsrt}

\title{Group Collaborative Learning for Co-Salient Object Detection}

\author{
Qi Fan$^{1,3}$\quad
Deng-Ping Fan$^{2,}$\thanks{Corresponding author: Deng-Ping Fan \emph{(dengpfan@gmail.com). 
}}\quad
Huazhu Fu$^{2}$\quad
Chi Keung Tang$^{1}$\quad
Ling Shao$^{2}$\quad
Yu-Wing Tai$^{1,3}$\\
$^1$ HKUST \quad
$^2$ Inception Institute of AI (IIAI) \quad
$^3$ Kwai Inc.\\
{\tt \small \url{https://github.com/fanq15/GCoNet}}
}

\maketitle

\begin{abstract}


\vspace{-10pt}
We present a novel group collaborative learning framework (\textbf{\ourmodel}) capable of detecting co-salient objects in real time (16ms), by simultaneously mining consensus representations at  group level based on the two necessary criteria:
\textbf{1) intra-group compactness} to better formulate the consistency among co-salient objects by capturing their inherent shared attributes using our novel group affinity module; 
\textbf{2) inter-group separability} to effectively suppress the influence of noisy objects on the output by introducing our new group collaborating module conditioning the inconsistent consensus. 
To learn a better embedding space without extra computational overhead, we explicitly employ auxiliary classification supervision.
Extensive experiments on three challenging benchmarks, i.e., CoCA, CoSOD3k, and Cosal2015,  demonstrate that our simple \ourmodel~outperforms 10 cutting-edge models and achieves the new state-of-the-art. We demonstrate this paper's new technical contributions on
a number of important downstream computer vision applications including content aware co-segmentation, co-localization based automatic thumbnails, etc. 

\end{abstract}

\section{Introduction}

Co-salient object detection (CoSOD) targets at detecting common salient objects sharing the same attributes given a group of relevant images. 
CoSOD is more challenging than the standard salient object detection (SOD) task,
because 
CoSOD needs to distinguish co-occurring objects across multiple images~\cite{deng2020re} in presence of other objects.
That is, both intra-class compactness and inter-class separability should be simultaneously maximized.
%
With this favorable feature CoSOD is thus often employed as a
pre-processing step for various computer vision tasks, such as image retrieval~\cite{cheng2014salientshape}, image quality assessment~\cite{wang2019no}, collection-based crops~\cite{jacobs2010cosaliency}, co-segmentation~\cite{wang2016higher,hsu2019deepco3}, semantic segmentation~\cite{zeng2019joint}, image surveillance~\cite{gao2020trustful}, video analysis~\cite{jerripothula2018efficient}, video co-localization~\cite{jerripothula2016cats}, \etc.   

%

\begin{figure}[!t]
\centering
\includegraphics[width=.85\linewidth]{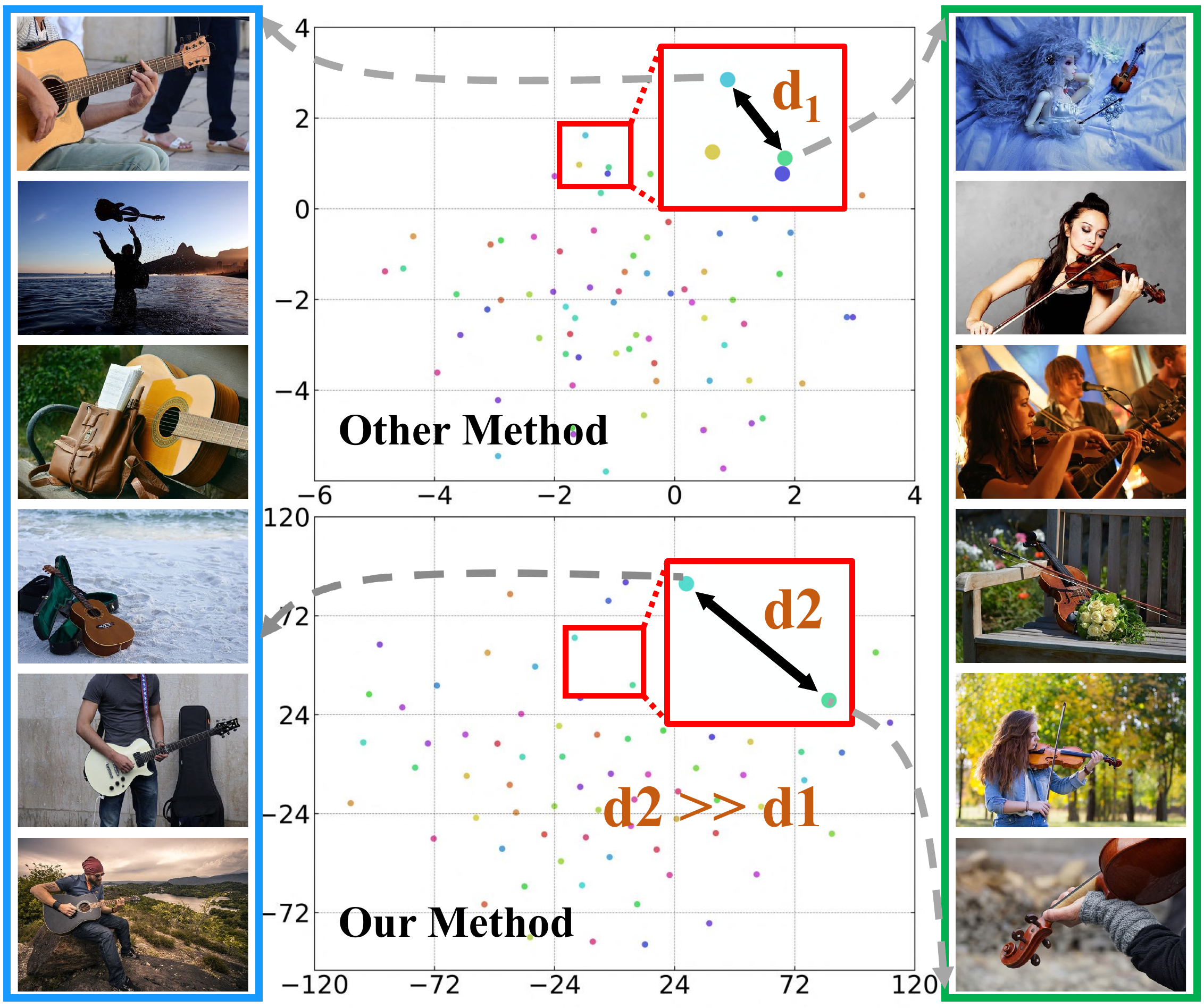}
\caption{\textbf{t-SNE~\cite{maaten2008visualizing} visualization of consensuses}, where each point represents one consensus of an image group. Highlighted here are two similar but different groups (\textit{guitar} \& \textit{violin}) to illustrate the effectiveness of \ourmodel.
The consensus strategy in traditional CoSOD model tends to cluster consensuses together even they belong to different groups, resulting in ambiguous co-saliency detection. In contrast, our consensus strategy with effective inter-group constraint enables higher
diversity with a very large group variance ($d_2 \gg d_1$) and thus better inter-group separability.}
\label{fig:teaser}
\end{figure}

Previous works attempt to leverage the consistency among relevant images to facilitate CoSOD \textit{within an image group} by exploring different shared cues~\cite{li2011co,fu2013cluster,cao2014self} or semantic connections~\cite{zhang2016co,han2017unified,hsu2018unsupervised}. 
Some of them~\cite{zou2013segmentation,cheng2014global} use  predicted saliency maps by computing various inter-image cues to discover  co-salient objects. Other works~\cite{deng2020re,zhang2020gradient} exploit a unified network to jointly optimize  co-saliency information and saliency maps. 

Despite their promising results, most current models only extract their CoSOD representations in an individual group, which introduces a number of limitations. First, images from the same group  contain  similar foregrounds (\ie, co-salient objects) only provide positive relations while lacking the negative relations between different objects. Training the model  only using positive pairs may lead to overfitting and result in ambiguous results for outlier images. Moreover, the  number of images in a group is typically limited (20 to 40 images for most CoSOD datasets), so using a single group cannot  provide enough information for learning a discriminative representation. Finally, individual groups also fall short in offering high-level semantic information, which is necessary for distinguishing noisy objects during inference in complex real-world scenarios.


To address the above issues, we propose a novel group collaborative learning framework (\textbf{\ourmodel}) to mine the semantic correlation between \textit{different image groups}. 
The proposed \ourmodel~consists of three important components:  group affinity module (GAM), group collaborating module (GCM) and auxiliary classification module (ACM), which simultaneously learn the \textbf{intra-group compactness} and \textbf{inter-group separability}. 
The GAM makes the network learn the consensus feature within the same image group, while the GCM discriminates target attributes between different groups, thus enabling the model to be trained on the existing large-scale SOD datasets.\footnote{Note that the existing CoSOD datasets altogether contain about 6k images, while there are more than 12 SOD datasets, containing about 60k images. It may partially alleviate the insufficient training data issue in co-salient object detection.}
We further improve the feature representation at a global semantic level through our ACM on each image to learn a better embedding space. 
In summary, our contributions are:
\begin{itemize}
    \item We introduce a novel group collaborative learning strategy to address the CoSOD problem, and validate its effectiveness with extensive ablation studies. 
    
    \item We design a novel unified Group Collaborative Learning Network (\textbf{\ourmodel}) for CoSOD by simultaneously considering intra-group compactness and inter-group separability to mine the consensus representation.
    
    \item Our group affinity module (GAM) and group collaborating module (GCM) collaborate with each other to achieve better intra- and inter-group collaborative learning. The auxiliary classification module (ACM) further promotes learning at a global semantic level.
    
    \item Extensive experiments on three challenging CoSOD benchmarks, \ie, CoCA, CoSOD3k, and Cosal2015, show that our {\ourmodel}  achieves the new state-of-the-art. Furthermore, we present two downstream applications based on our technical contributions, \ie, co-segmentation and co-localization. 
\end{itemize}

\section{Related Work}

The traditional salient object detection task~\cite{qin2019basnet,liu2019simple,zhao2019egnet,GaoEccv20Sal100K,zhao2019optimizing} targets at directly segmenting salient object in each image separately, while CoSOD aims to segment the common salient objects across several relevant images. Previous works mainly exploit inter-image cues to detect co-salient objects. Early CoSOD methods explore the inter-image correspondence between image-pairs~\cite{li2011co,chen2010preattentive} or a group of relevant images~\cite{cao2014co} based on shallow handcrafted descriptors~\cite{jerripothula2016cats,chang2011co}. They employ different approaches to mine the inter-image relationships using constraints or heuristic characteristics.
Several studies attempt to capture the inter-image constraints by employing an efficient manifold ranking scheme~\cite{li2014efficient} to obtain guided saliency maps, or using a global
association constraint with clustering~\cite{fu2013cluster}, or translational alignment~\cite{jacobs2010cosaliency}.
Other works attempt to formulate the semantic attributes shared among images in a group from the high-level features in the heuristic characteristics, using  multiple saliency cues and self-adaptive weights~\cite{cao2014self}, regional histograms and constrasts~\cite{liu2013co}, metric learning by optimizing a new objective function~\cite{han2017unified}, or pairwise similarity ranking and linear programming~\cite{li2013co}.

Recently deep-based models simultaneously explore the intra- and inter-image consistency in a supervised manner with different approaches, such as graph convolution networks (GCN)~\cite{jiang2019unified,jiang2019multiple,zhang2020adaptive}, self-learning methods~\cite{zhang2016co,zhang2015cosaliency}, inter-image co-attention with PCA projection~\cite{deng2020re} or recurrent units~\cite{li2019detecting}, correlation techniques~\cite{icnet2020nips}, quality measurement~\cite{jerripothula2018quality}, or co-clustering~\cite{yao2017revisiting}. Some methods exploit multi-task learning to simultaneously optimize the co-saliency detection and co-segmentation~\cite{tsai2018image} or co-peak search~\cite{hsu2019deepco3}. Other works explore hierachical features from multi-scale~\cite{zhang2019co}, multi-stage~\cite{li2018deep}, or multi-layer~\cite{ren2020co} features. Another notable research line is to explore  group-wise semantic representation (consensus) which is used to detect co-salient regions for each image. There are different methods to capture the discriminative semantic representation, such as group attentional semantic aggregation~\cite{coadnet2020nips}, gradient feedback~\cite{zhang2020gradient}, co-category association ~\cite{wang2019robust}, united fully convolutional network~\cite{wei2019deep,wei2017group}, or integrated multilayer graph~\cite{jeong2018co}.  Methods are proposed to solve the CoSOD problem in a semi-supervised~\cite{zheng2018feature} or unsupervised manner~\cite{hsu2018unsupervised,zhang2017supervision,hsu2018co,li2019co}, and studies~\cite{yu2018co,song2019easy} are availalbe on co-saliency detection from a single image.

Previous works have focused on  intra-group (intra- and inter-image) cues for capturing common attributes of co-salient objects. The inter-group information
has received less attention, although CODW~\cite{zhang2016detection} focuses on \textit{visually similar} neighbor.
Recently Zhang \textit{et al.}~\cite{zhang2020gradient} utilized a jigsaw training to \textit{implicitly} exploit other images to facilitate group training. But their model still targets intra-group learning. Our method differs from existing models in the exploration of inter-group relations for discriminating feature learning at a group level \textit{explicitly and semantically}.

\begin{figure*}[!t]
\centering
\includegraphics[width=.85\linewidth]{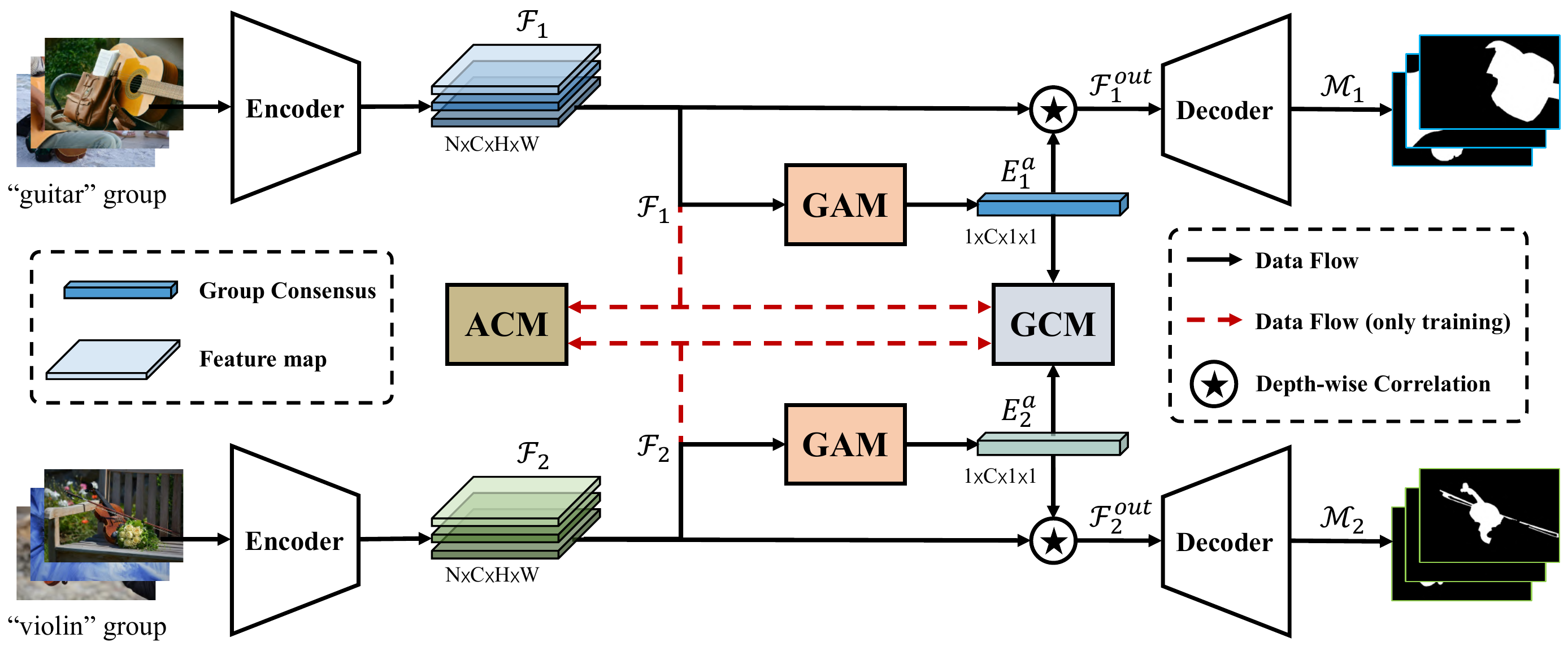}
\caption{\textbf{Pipeline of the proposed Group Collaborative Learning Network (\ourmodel).} Images in two groups are first processed by a weight-shared encoder. Then we employ the group affinity module (GAM, see Figure~\ref{fig:gam} for more details) to conduct intra-group collaborative learning for each group to generate a consensus, which is collaborated with the original feature maps to segment co-salient objects using the decoder. In addition, the original feature maps and consensuses of both groups are fed to the group collaborating module (GCM, see Figure~\ref{fig:gcm}) to conduct the inter-group collaborative learning. Moreover an auxiliary classification module (ACM) is applied to obtain the high-level semantic representation. The GCM and ACM are only used for training and are removed at inference.}
\label{fig:network}
\end{figure*}

\section{Group Collaborative Learning Network}

\subsection{Architecture Overview}


Given a group of $N$ relevant images 
$\{I_1, I_2, ..., I_n\}$
containing common salient objects of a certain class, CoSOD aims to detect them simultaneously and output the co-saliency maps. 
Unlike  existing CoSOD methods which only depend on the information within the image group, we propose a novel group collaborative learning network (\ourmodel) to mine the consensus representations at both  intra- and inter-group level. 

Figure~\ref{fig:network} illustrates the flowchart of our \ourmodel. First, an encoder network is used to extract feature maps $\mathcal{F}_1=\{F_{1,n}\}_{n=1}^N, \mathcal{F}_2=\{F_{2,n}\}_{n=1}^N \in \mathbb{R}^{N \times C \times H \times W}$ for two image groups, where $C$ is the channel number and $H \times W$ is the spatial size. 
Then, a group affinity module (GAM) is used to combine all single-image features  to distill the consensus feature $\textit{E}_1^a, \textit{E}_2^a \in \mathbb{R}^{1 \times C \times 1 \times 1}$ from $\mathcal{F}_1, \mathcal{F}_2$, representing the common attributes of the co-salient objects for each group.  Simultaneously, a group collaborating module (GCM) is applied to enhance the  image representation for discriminating the target attributes between  different image groups. Finally, we further improve the high-level semantic representation of images using an auxiliary classification module (ACM) to learn a better embedding space. The resulting collaborative features are then fed to a decoder network to produce the co-saliency maps $\mathcal{M}_1,\mathcal{M}_2$.



\begin{figure*}[!t]
\centering
\includegraphics[width=.85\linewidth]{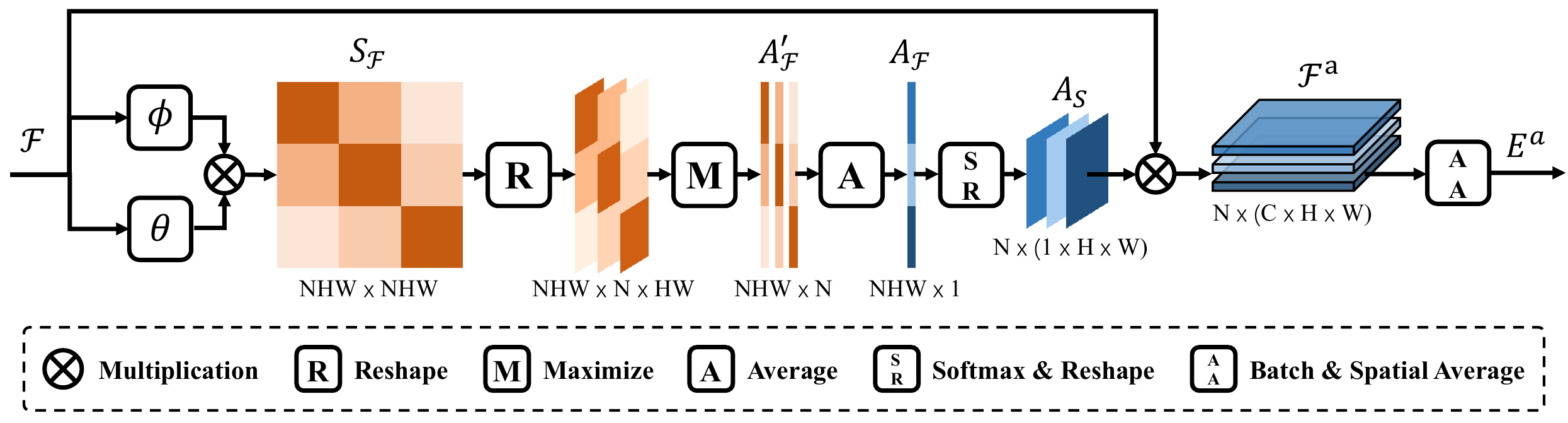}
\caption{\textbf{Group Affinity Module.} We first exploit the affinity attention to generate the attention maps for the input features by collaborating all images in group. Subsequently, the maps are multiplied with the input features to generate the consensus for the group. Then the obtained consensus is used to coordinate the original feature maps and is also fed to the GCM for inter-group collaborative learning.}
\label{fig:gam}
\end{figure*}

\subsection{Group Affinity Module}
\label{section:3.2}

Intuitively, common objects from the same class always share some similarity in appearance and have high similarity in features, which have been widely employed in many tasks. Inspired by self-supervised video tracking methods~\cite{vondrick2018tracking,lai2019self,wang2019learning,lai2020mast}, which propagate the segmentation masks of target objects based on the pixel-wise correspondences between two adjacent frames, we extend this idea to the CoSOD task by computing the global affinity among all images in a group. 

For any two image features $\{F_{1,n}, F_{1,m}\} \in \mathcal{F}_1$ \footnote{All  analyses in section~\ref{section:3.2}  on $\mathcal{F}_1$ can be applied to 
$\mathcal{F}_2$. We omit the group subscript for notation simplicity, \ie, we use $F_n$ to represent $F_{1,n}$.} and without losing generality we drop the group subscript, we can use the inner product to compute their pixel-wise correlations:
\begin{equation}
	S_{(n,m)} =  \theta (F_n)^T \phi(F_m)\label{equation:1},
\end{equation}
where $\theta, \phi$ are linear embedding functions. The affinity map $S_{(n,m)} \in \mathbb{R}^{HW \times HW}$ efficiently captures the commonality of co-salient objects in the image pair $(n,m)$. Then we can generate $F_n$'s affinity map $A_{n \leftarrow m} \in \mathbb{R}^{HW \times 1}$ by finding the maxima for each of $F_n$'s pixel conditioned on $F_m$ which alleviates the influence of noisy correlation values in the map.

Similarly, we can extend the local affinity of two images to the global affinity of all images in the group. 
Specifically, we compute the affinity map $S_{\mathcal{F}} \in \mathbb{R}^{NHW \times NHW}$ between all image features $\mathcal{F}$ using Eq.~\ref{equation:1}. Then, we find the maxima for each image $A_{\mathcal{F}}' \in \mathbb{R}^{NHW \times N}$ from $S_{\mathcal{F}}$, and average all the maxima of $N$ images to generate the global affinity attention map $A_{\mathcal{F}} \in \mathbb{R}^{NHW \times 1}$. 
In this way, the affinity attention map is globally optimized on all images thus alleviating the influence of occasional co-occurring bias. Then, we use a softmax operation to normalize $A_{\mathcal{F}}$ and reshape it to generate the attention map $A_S \in \mathbb{R}^{N \times (1 \times H \times W)}$. We multiply $A_S$ with the original feature $\mathcal{F}$ to produce the attention feature maps $\mathcal{F}^a \in \mathbb{R}^{N \times C \times H \times W}$.
Finally, all the attention feature maps $\mathcal{F}^a$ for the whole group are used to produce the attention consensus $\textit{E}^a$ by average pooling along both the batch and spatial dimensions, as shown in Figure~\ref{fig:gam}. 

The global affinity module focuses on capturing the commonality among co-salient objects within the same group and therefore improves the intra-group compactness of the consensus representation. Such \textit{intra-group compactness} alleviates the disturbance of co-occurring noise and enables the model to concentrate on the co-salient regions. This allows the shared attributes of co-salient objects to be better captured and therefore results in better consensus representation.
The obtained attention consensus $\textit{E}^a$ is combined with the original feature maps $\mathcal{F}$ through depth-wise correlation~\cite{li2019siamrpn++,fan2020few} to achieve efficient information association. The resulting feature maps $\mathcal{F}^{out}$ are fed to the decoder to predict co-saliency maps $\mathcal{M}_n$ for each image. The loss function is:
\begin{equation}
	\mathcal{L}_{\text{sal}}=\frac{1}{N} \sum_n^N \mathcal{L}_{\text{siou}}(\mathcal{M}_n,\mathcal{G}_n),
\end{equation}
where $\mathcal{L}_{\text{siou}}$ is the soft IoU loss~\cite{qin2019basnet,li2018interactive} and $\mathcal{G}_n$ denotes the ground-truth label for each image in the group.


\begin{figure}[!t]
\centering
\includegraphics[width=1.0\linewidth]{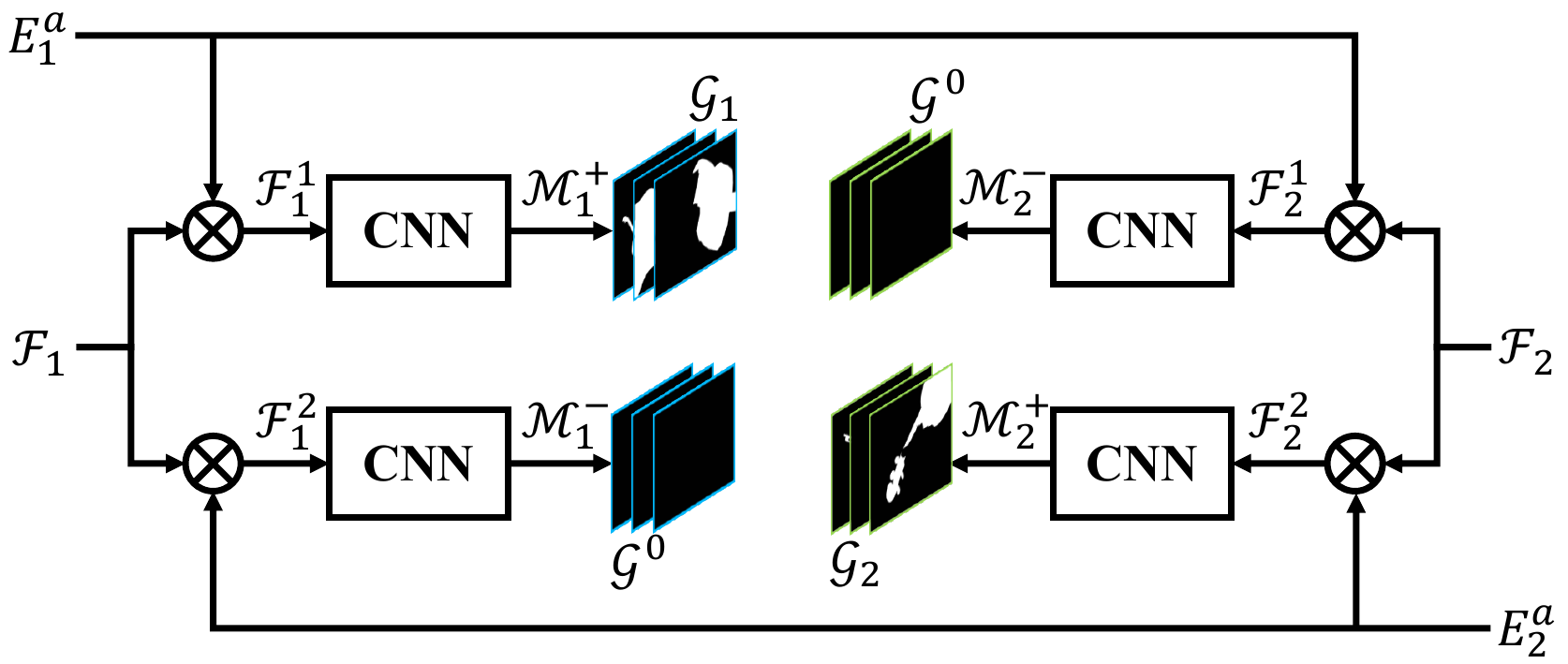}
\caption{\textbf{Group collaborating module.} The original feature maps and consensuses of both groups are fed to the GCM. The predicted output conditioned on the consistent feature and consensus (from the same group) is supervised with the available ground-truth labels. Otherwise, it is supervised by the all-zero maps.}
\label{fig:gcm}
\end{figure}

\subsection{Group collaborating module (GCM)}

Most CoSOD methods tend to focus on the intra-group compactness of the consensus, but the \textit{inter-group separability} is equally crucial for distinguishing distracting objects, especially when processing complex images with more than one salient objects. 
To enhance the discriminative representations between different groups, we propose a simple but effective module, \ie, the GCM, by learning to encode the inter-group separability.

Given two image groups with the corresponding features $\{\mathcal{F}_1, \mathcal{F}_2\}$ and attention consensus $\{\textit{E}_1^a, \textit{E}_2^a\}$ obtained from the GAM, we apply an intra- and inter-group cross-multiplication. Specifically, the intra-group multiplication deals with the features and their consensus: $\mathcal{F}_1^1 = \mathcal{F}_1 \cdot \textit{E}_1^a$ and $\mathcal{F}_2^2= \mathcal{F}_2 \cdot \textit{E}_2^a$ for the intra-group collaboration, while the inter-group multiplication acts on the features and consensus of different groups, \ie, $\mathcal{F}_1^2= \mathcal{F}_1 \cdot \textit{E}_2^a$ and $\mathcal{F}_2^1= \mathcal{F}_2 \cdot \textit{E}_1^a$, to express the inter-group interaction.
The intra-group representation $\mathcal{F}^+ = \{\mathcal{F}_1^1, \mathcal{F}_2^2\}$ is exploited to predict the co-saliency maps, and the inter-group representation $\mathcal{F}^- = \{\mathcal{F}_1^2, \mathcal{F}_2^1\}$ is employed to provide a consensus with group separability.
Specifically, we feed $\{\mathcal{F}^+, \mathcal{F}^-\}$ to a small convolutional network with an upsampling layer and produce the saliency map $\{\mathcal{M}^+, \mathcal{M}^-\}$\footnote{$\mathcal{M}^+ = \{\mathcal{M}_1^+, \mathcal{M}_2^+\}$ and $\mathcal{M}^- = \{\mathcal{M}_1^-, \mathcal{M}_2^-\}$.}
with different supervision signals: we use  ground-truth labels to supervise  $\mathcal{F}^+$, while all-zero maps are used for $\mathcal{F}^-$. The loss function is: 
\begin{equation}
	\mathcal{L}_{\text{ctm}} = \frac{1}{N} \sum_n^N \mathcal{L}_{\text{FL}}(<\mathcal{M}_n^+, \mathcal{M}_n^->, <\mathcal{G}_n, \mathcal{G}_n^0>),
\end{equation}
where $\mathcal{L}_{\text{FL}}$ is the focal loss~\cite{lin2017focal}, $\mathcal{G}_n$ is the ground-truth, $\mathcal{G}_n^0$ is the all-zero map and $<\cdot>$ denotes the concatenation operation.

\begin{table*}
\begin{center}
\footnotesize
\renewcommand{\tabcolsep}{1.8mm}
\caption{\textbf{Quantitative ablation studies} of our \ourmodel~on the effectiveness of the GAM (group affinity module), GCM (group collaborating module), ACM (auxiliary classification module) and their combinations. }
\label{table:ablation}
\begin{tabular}{c|ccc||cccc|cccc|cccc}
\hline
& \multicolumn{3}{c||}{Modules}  & \multicolumn{4}{|c|}{CoCA~\cite{zhang2020gradient}} & \multicolumn{4}{|c|}{CoSOD3k~\cite{deng2020re}} & \multicolumn{4}{|c}{Cosal2015~\cite{zhang2016detection}} \\
ID &  GAM & GCM & ACM & $E_\phi^\text{ max} \uparrow$ & $S_\alpha \uparrow$ & $F_\beta^\text{ max} \uparrow$ & $\epsilon \downarrow$ & $E_\phi^\text{ max} \uparrow$ & $S_\alpha \uparrow$ & $F_\beta^\text{ max} \uparrow$ & $\epsilon \downarrow$ & $E_\phi^\text{ max} \uparrow$ & $S_\alpha \uparrow$ & $F_\beta^\text{ max} \uparrow$ & $\epsilon \downarrow$ \\
\hline
1 &  &  &                                & 0.618 & 0.591 & 0.419 & 0.190 & 0.811 & 0.764 & 0.721 & 0.108 & 0.862 & 0.818 & 0.800 & 0.087 \\
2 & \checkmark &  &                      & 0.663 & 0.605 & 0.442 & 0.160 & 0.823 & 0.772 & 0.736 & 0.099 & 0.873 & 0.825 & 0.815 & 0.079 \\
3 &  & \checkmark &                      & 0.666 & 0.616 & 0.452 & 0.156 & 0.839 & 0.788 & 0.748 & 0.087 & 0.877 & 0.834 & 0.823 & 0.074 \\
4 &  &  & \checkmark                     & 0.651 & 0.606 & 0.442 & 0.167 & 0.829 & 0.779 & 0.737 & 0.094 & 0.875 & 0.832 & 0.820 & 0.076 \\
5 & \checkmark & \checkmark &            & 0.719 & 0.650 & 0.504 & 0.126 & 0.850 & 0.798 & 0.766 & 0.078 & 0.884 & 0.842 & 0.837 & 0.070 \\
\hline
& \checkmark & \checkmark & \checkmark & \textbf{0.760} & \textbf{0.673} & \textbf{0.544}  & \textbf{0.105} & \textbf{0.860} & \textbf{0.802} & \textbf{0.777}  & \textbf{0.071} & \textbf{0.888} & \textbf{0.845} & \textbf{0.847}  & \textbf{0.068} \\
\hline
\end{tabular}
\end{center}
\vspace{-0.1in}
\end{table*}

Our GCM thus encourages the consensus to distinguish different groups with high
inter-group separability to identify distractors in complex environment. Another advantage is that this module enables the model to be trained on the existing SOD datasets, whose images typically contain only one dominating object. We can discard this module during inference without introducing additional computational overhead. 

\subsection{Auxiliary Classification Module (ACM)}


To obtain more discriminative features for consensus, we also introduce an ACM to facilitate high-level semantic representation learning. Specifically, we add a classification predictor with a global average pooling layer and one fully connected layer to the backbone to classify $F_n$ to the corresponding class $\mathcal{Y}_n$. In the Euclidean feature space, the classification supervision can separate classes by introducing a large margin, and cluster samples belonging to the same class. Therefore, it enables the model to generate more representative features and benefits the consensus learning for intra-group compactness and inter-group separability. The loss function is:
\begin{equation}
	\mathcal{L}_{\text{cls}} = \mathcal{L}_{\text{ce}}(\mathcal{Y}_n, \hat{\mathcal{Y}_n}),
\end{equation}
where $\mathcal{L}_{\text{ce}}$ is the cross-entropy loss and $\hat{\mathcal{Y}_n}$ is the ground-truth class label.

\subsection{End-to-end Training}

During training, the GAM, GCM, and ACM are jointly trained with the backbone in an end-to-end manner. The whole framework is optimized by integrating all the aforementioned loss functions:
\begin{equation}
    \mathcal{L} = \lambda_1\mathcal{L}_{\text{sal}} + \lambda_2\mathcal{L}_{\text{ctm}} + \lambda_3\mathcal{L}_{\text{cls}},
\end{equation}
where $\lambda_1$, $\lambda_2$, and $\lambda_3$ are hyperparameter weights to balance the loss functions.

\section{Experiments}
\subsection{Implementation Details}


We use VGG-16~\cite{simonyan2014very} with Feature Pyramid Network (FPN)~\cite{lin2017feature} as our backbone. 
For fair comparison, we follow GICD~\cite{zhang2020gradient} and use the DUTS~\cite{wang2017learning} dataset as our training set. The group labels derived from  GICD~\cite{zhang2020gradient} are used to group the images during training. In each training episode, we randomly pick two different groups with 16 samples\footnote{Due to limited computing resource. The larger the better.} in each  group to train the network. 
The images are all resized to 224x224 for training and testing, and the output saliency maps are resized to the original size for evaluation. The network is trained over $50$ epochs in total with the Adam optimizer. The initial learning rate is set to $10e-4$, $\beta_1=0.9$ and $\beta_2=0.99$. 
The whole training takes around four hours and the inference speed on the image pair groups\footnote{CoSOD task works for image groups. Therefore we use the basic image pair group to evaluate the speed rather than the single image.} is $16$ $ms$. The platform for training and inference is equipped with $56$ Intel(R) Xeon(R) CPU E5-2680 v4 @ 2.40GHz and a Nvidia GeForce GTX 1080Ti.


\subsection{Evaluation Datasets and  Metrics}
We employ three challenging datasets for evaluation:
CoCA~\cite{zhang2020gradient}, CoSOD3k~\cite{fan2020taking}, and Cosal2015~\cite{zhang2016detection}. The last is a large dataset widely used in the evaluation of CoSOD methods. The first two were recently proposed for  challenging real-world co-saliency evaluation, with the images usually containing multiple common and non-common objects against a complex background. 
Following the advice of recent large-scale benchmark work~\cite{fan2020taking}, we do not use iCoseg~\cite{batra2010icoseg} and MSRC~\cite{winn2005object} for evaluation, because tbey usually provide only one salient object in an image and are not very suitable for evaluating CoSOD models.
%
We use maximum E-measure $E_\phi^\text{max}$~\cite{fan2018enhanced},  S-measure 
$S_\alpha$~\cite{fan2017structure},  maximum F-measure $F_\beta^\text{max}$~\cite{achanta2009frequency}, and mean absolute error (MAE) $\epsilon$~\cite{cheng2013efficient} to evaluate methods in our experiments. Evaluation toolbox: \url{https://github.com/DengPingFan/CoSODToolbox}. 

\begin{table*}
\begin{center}
\footnotesize
\renewcommand{\arraystretch}{}
\renewcommand{\tabcolsep}{1.2mm}
\caption{\textbf{Quantitative comparison results} between our \ourmodel~and other methods. ``$\uparrow$'' (``$\downarrow$'') means that the higher (lower) is better. Co = CoSOD models, Sin = Single-SOD models.
The symbol $*$ denotes traditional CoSOD algorithms.  Online benchmark has been made publicly available at: \url{http://dpfan.net/cosod3k}.
}
\label{table:coco}
\begin{tabular}{r||r|c|cccc|cccc|cccc}
\hline
& & &\multicolumn{4}{|c|}{CoCA~\cite{zhang2020gradient}} & \multicolumn{4}{|c|}{CoSOD3k~\cite{deng2020re}} & \multicolumn{4}{|c}{Cosal2015~\cite{zhang2016detection}} \\
Method & Pub. \& Year & Type & $E_\phi^\text{ max} \uparrow$ & $S_\alpha \uparrow$ & $F_\beta^\text{ max} \uparrow$ & $\epsilon \downarrow$ & $E_\phi^\text{ max} \uparrow$ & $S_\alpha \uparrow$ & $F_\beta^\text{ max} \uparrow$ & $\epsilon \downarrow$ & $E_\phi^\text{ max} \uparrow$ & $S_\alpha \uparrow$ & $F_\beta^\text{ max} \uparrow$ & $\epsilon \downarrow$ \\
\hline
CBCD*~\cite{fu2013cluster} & TIP 2013 &	 Co & 0.641 & 0.523 & 0.313 & 0.180 & 0.637 & 0.528 & 0.466 & 0.228 & 0.656 & 0.544 & 0.532 & 0.233 \\
GWD~\cite{wei2017group} & IJCAI 2017 & Co & 0.701 & 0.602 & 0.408 & 0.166 & 0.777 & 0.716 & 0.649 & 0.147 & 0.802 & 0.744 & 0.706 & 0.148 \\
RCAN~\cite{li2019detecting} & IJCAI 2019 & Co & 0.702 & 0.616 & 0.422 & 0.160 & 0.808 & 0.744 & 0.688 & 0.130 & 0.842 & 0.779 & 0.764 & 0.126 \\
CSMG~\cite{zhang2019co} & CVPR 2019 & Co & 0.733 & 0.627 & 0.499 & 0.114 & 0.804 & 0.711 & 0.709 & 0.157 & 0.842 & 0.774 & 0.784 & 0.130 \\
BASNet~\cite{qin2019basnet} & CVPR 2019 & Sin & 0.644 & 0.592 & 0.408 & 0.195 & 0.804 & 0.771 & 0.720 & 0.114 & 0.849 & 0.822 & 0.791 & 0.096 \\
PoolNet~\cite{liu2019simple} & CVPR 2019 & Sin & 0.640 & 0.602 & 0.404 & 0.177 & 0.799 & 0.771 & 0.709 & 0.113 & 0.848 & 0.823 & 0.785 & 0.094 \\
EGNet~\cite{zhao2019egnet} & ICCV 2019& Sin & 0.648 & 0.603 & 0.404 & 0.178 & 0.793 & 0.762 & 0.702 & 0.119 & 0.843 & 0.818 & 0.786 & 0.099 \\
SCRN~\cite{wu2019stacked} & ICCV 2019 & Sin & 0.642 & 0.612 & 0.413 & 0.164 & 0.805 & 0.771 & 0.716 & 0.113 & 0.850 & 0.817 & 0.783 & 0.098 \\
GICD~\cite{zhang2020gradient} & ECCV 2020 & Co & 0.715 & 0.658 & 0.513 & 0.126 & 0.848 & 0.797 & 0.770 & 0.079 & \textbf{0.887} & 0.844 & 0.844 & 0.071 \\
CoEGNet~\cite{deng2020re} & TPAMI 2021 &	Co  & 0.717 & 0.612 & 0.493 & 0.106 & 0.825 & 0.762 & 0.736 & 0.092 & 0.882 & 0.836 & 0.832 & 0.077 \\
\hline

\hline
\textbf{\ourmodel~(Ours)}	& CVPR 2021 &	Co	  & \textbf{0.760} & \textbf{0.673} & \textbf{0.544} & \textbf{0.105} & \textbf{0.860} & \textbf{0.802} & \textbf{0.777} & \textbf{0.071} & \textbf{0.887} & \textbf{0.845} & \textbf{0.847} & \textbf{0.068} \\
\hline
\end{tabular}
\label{table:main}
\end{center}
\vspace{-0.1in}
\end{table*}

\begin{figure*}
\begin{center}
\includegraphics[width=.95\linewidth]{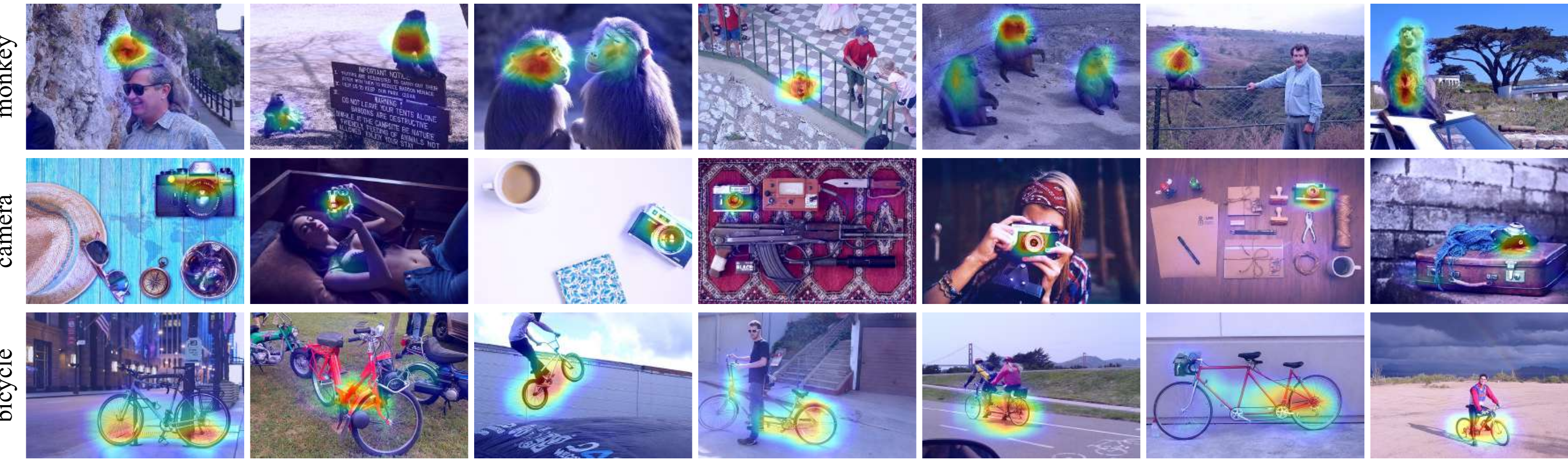}
\end{center}
\vspace{-5pt}
   \caption{\textbf{Visualization of affinity attention maps} learned by GAM using intra-group collaborative learning across all images in each group. Masks are sensitive to co-salient regions with shared attributes, which benefits the consensus representation learning.}
\label{fig:GAM}
\end{figure*}

\subsection{Ablation Studies}

In this section, we study the effectiveness of each component in our approach (Table~\ref{table:ablation}) and investigate how they contribute to a good consensus feature.

\noindent{\bf Effectiveness of GAM.~}
The global co-attention module is a fundamental component of our model, which is designed to capture the common attributes of co-salient objects in an image group for better \textit{intra-group compactness}. Compared to the baseline model with only the vanilla consensus extracted by an average pooling operation, GAM improves the performance on all  metrics and datasets. To get a deeper understanding of our GAM module, we visualize the learned attention masks in Figure~\ref{fig:GAM}. We find that our global co-attention effectively alleviates the influence of co-occurring noise and focuses on co-salient regions in the image groups, \eg, in both the {\it monkey} and {\it bicycle} groups, there are some co-occurring {\it persons} in some images, but our GAM is not adversely influenced. The global view of GAM enables the most common objects to be detected, while the local pair-wise co-attention cannot distinguish them in the local view.

\noindent{\bf Effectiveness of GCM.~}
The group collaborating module is designed to enable the consensus \textit{inter-group separability} to distinguish distracting objects from non-common objects. 
After equip  the model with GCM, significant performance improvement (ID-1~versus~ID-3) is obtained in Table~\ref{table:ablation} especially on the challenging CoCA~\cite{zhang2020gradient} dataset whose images usually contain multiple uncommon and common objects.
To investigate the consensus characteristics when the model is trained with the GCM, we visualize the consensus using t-SNE~\cite{maaten2008visualizing} on the CoCA dataset, and compare with the vanilla consensus without the GCM.
As shown in Figure~\ref{fig:teaser}, the vanilla consensuses (top: other method) tend to cluster together, even if they belong to different groups, resulting in ambiguous co-saliency detection, especially for objects belonging to similar but different groups. 
In contrast, the consensuses trained with the GCM (bottom: our method) is more diverse with a higher group variance  ($d_2 \gg d_1$), for more effective inter-group separability. 

\noindent{\bf Effectiveness of ACM.~}
As shown in Table~\ref{table:ablation}, the classification module introduces better backbone features for the consensus with the auxiliary classification supervision. The ACM improves the baseline performance on all metrics and datasets. This cost-free improvement does not change the network architecture and does not introduce extra computational overhead, thus has substantial potential to other models and tasks to take advantage of the multi-task learning and more representative features.

\begin{figure*}[t!]
	\centering
	\includegraphics[width=.96\textwidth]{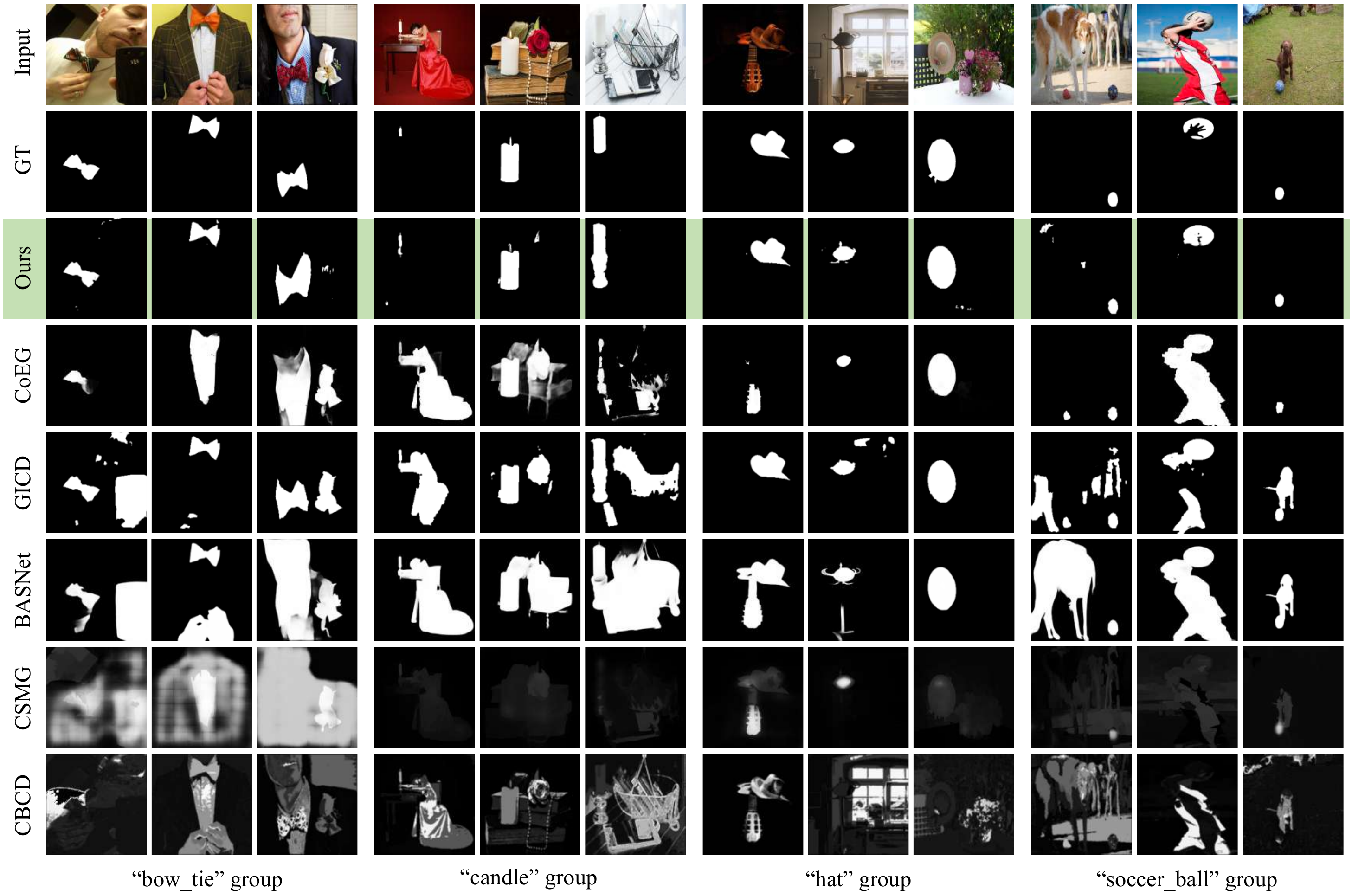}
	\caption{\textbf{Qualitative comparisons} of our \ourmodel~and other methods.}
	\label{fig:vis}
\end{figure*}

\subsection{Competing Methods}
Since not all CoSOD models have publicly released codes, we only compare our {\ourmodel}  with one representative traditional algorithm (CBCD) and five deep-based CoSOD models, including GWD~\cite{li2019group}, RCAN~\cite{li2019detecting}, CSMG~\cite{zhang2019co}, GICD~\cite{zhang2020gradient}, and 
CoEGNet~\cite{deng2020re}. Following the current state-of-the-art model~\cite{zhang2020gradient}, we also compare with four cutting-edge deep salient object detection (SOD)\footnote{SOD methods can also be directly applied to the CoSOD task.} models: BASNet~\cite{qin2019basnet}, PoolNet~\cite{liu2019simple}, EGNet~\cite{zhao2019egnet} and SCRN~\cite{wu2019stacked}. More complete leaderboard can be found in recent standard benchmark works~\cite{deng2020re,fan2020taking}.



\paragraph{Quantitative Results.~}
Table~\ref{table:main}
tabulates the quantitative results of our model and state-of-the-art methods. Our model outperforms all of them in all metrics, especially on the challenging CoCA and CoSOD3k datasets. Among these three datasets, CoCA is the most challenging, since the images typically  contain other multiple  objects in addition to the co-salient objects which are even smaller in size. Our model capitalizes on our better consensus  
and significantly outperforms other methods especially  the SOD methods which are trapped in distinguishing many distracting objects instead. 
CoSOD3k has similar attributes, and our model still performs much better than other models on this dataset. Cosal2015 is the easiest dataset because its images typically only contain one co-salient object, and therefore the SOD algorithms can easily handle this dataset. Our model cannot take full advantage of the better consensus on this dataset and the improvement is not as significant as on other datasets.

\paragraph{Qualitative Results.~} Figure~\ref{fig:vis} shows the saliency maps generated by different methods for qualitative comparison. In these difficult examples, each image contains other multiple objects in addition to the co-salient objects. As aforementioned, the SOD methods can only detect  salient objects and fail to distinguish co-salient objects due to their intrinsic limitation. The CoSOD methods perform better than the SOD methods owing to their consensus for distinguishing co-salient regions. However, limited by the their weak consensus, they are still unable to  handle the challenging cases. Our model introduces an effective consensus  through optimizing intra-group compactness and inter-group separability, and therefore performs much better on detecting co-salient objects.

\begin{figure*}
\begin{center}
\includegraphics[width=.985\linewidth]{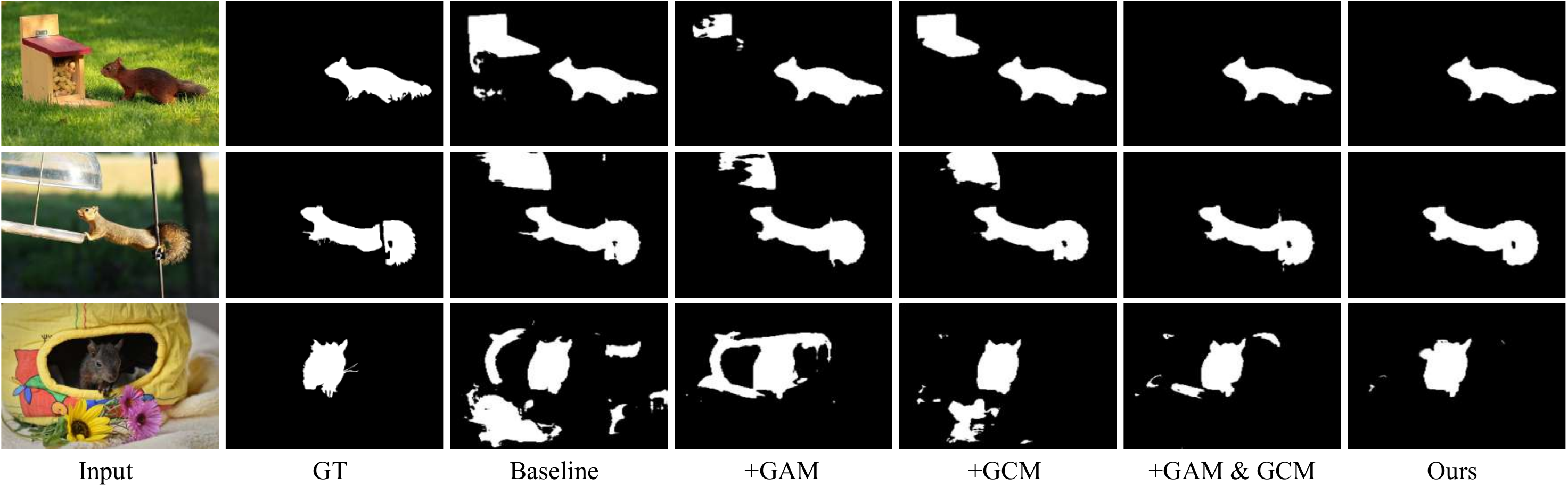}
\end{center}
\vspace{-15pt}
   \caption{\textbf{Qualitative ablation studies} of our \ourmodel~on different modules and their combinations. }
\label{fig:ablation}
\vspace{-0.2cm}
\end{figure*}

\begin{figure}
\begin{center}
\includegraphics[width=.96\linewidth]{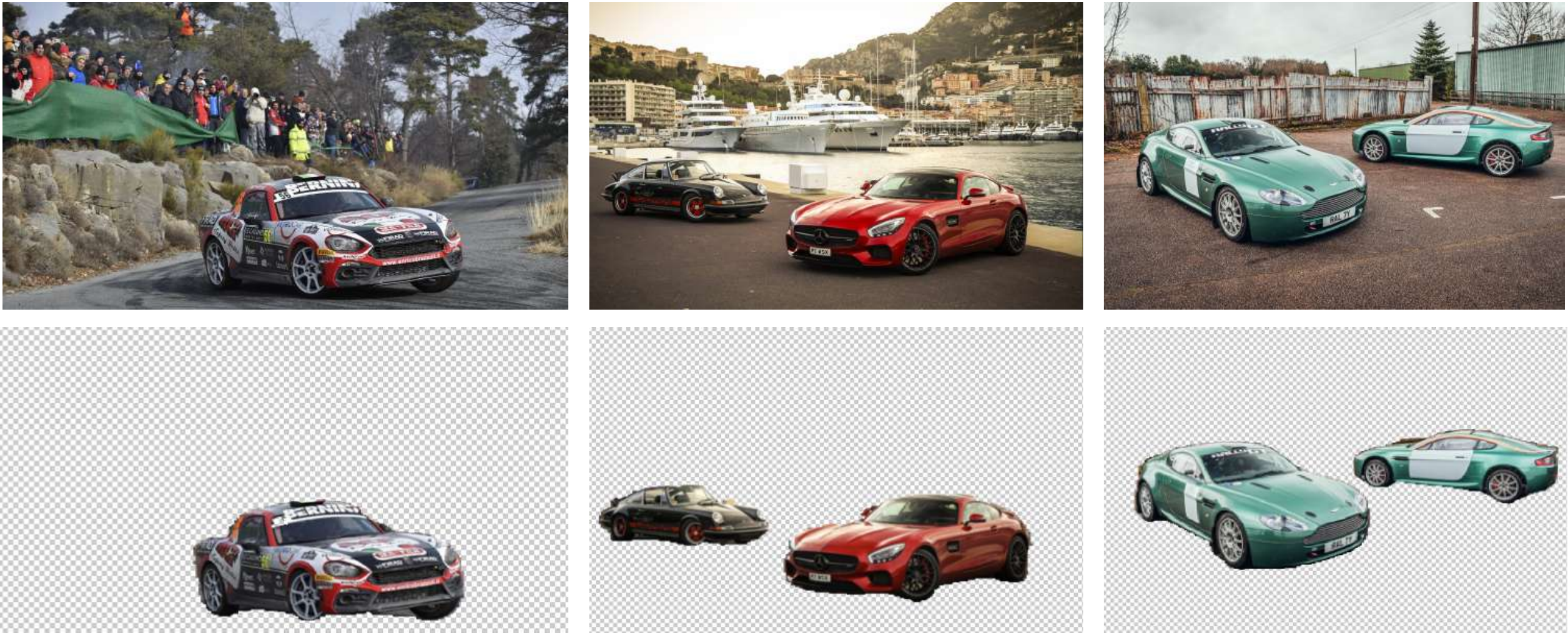}
\end{center}
\vspace{-10pt}
\caption{\textbf{Application 1.} Content aware object co-segmentation visual results (``GT car") obtained by our \ourmodel.}
\label{fig:app1}
\vspace{-10pt}
\end{figure}

\section{Discussion of Module Cooperation}
Our three modules are closely interdependent and mutually reinforced for improving co-saliency detection performance. Combining the GAM and GCM can significantly improve the performance compared to the individual modules. 
Without the GAM the vanilla consensus is not robust against 
noise caused by  uncommon objects and background, and the low-quality consensus cannot take full advantage of the GCM which heavily relies on the consensus for distinguishing different objects. On the other hand, although the consensus can capture  common attributes with the help of the GAM, it is difficult to distinguish different groups without the GCM especially for similar groups. Overall, the GAM produces better consensus with high intra-group compactness to detect co-saliency objects, while the GCM further endows the consensus with inter-group separability for better discriminative ability. Adding ACM, the consensus can benefit from more representative features leveraged by the multi-task learning.

Figure~\ref{fig:ablation} qualitatively analyse their cooperation. The baseline model detects uncommon objects, while the GAM and GCM can slightly ameliorate their adverse influence. When combining the GAM and GCM, the model can effectively capture co-salient objects with the ACM further boosting the co-salient object detection result.

\begin{figure}
\begin{center}
\includegraphics[width=.96\linewidth]{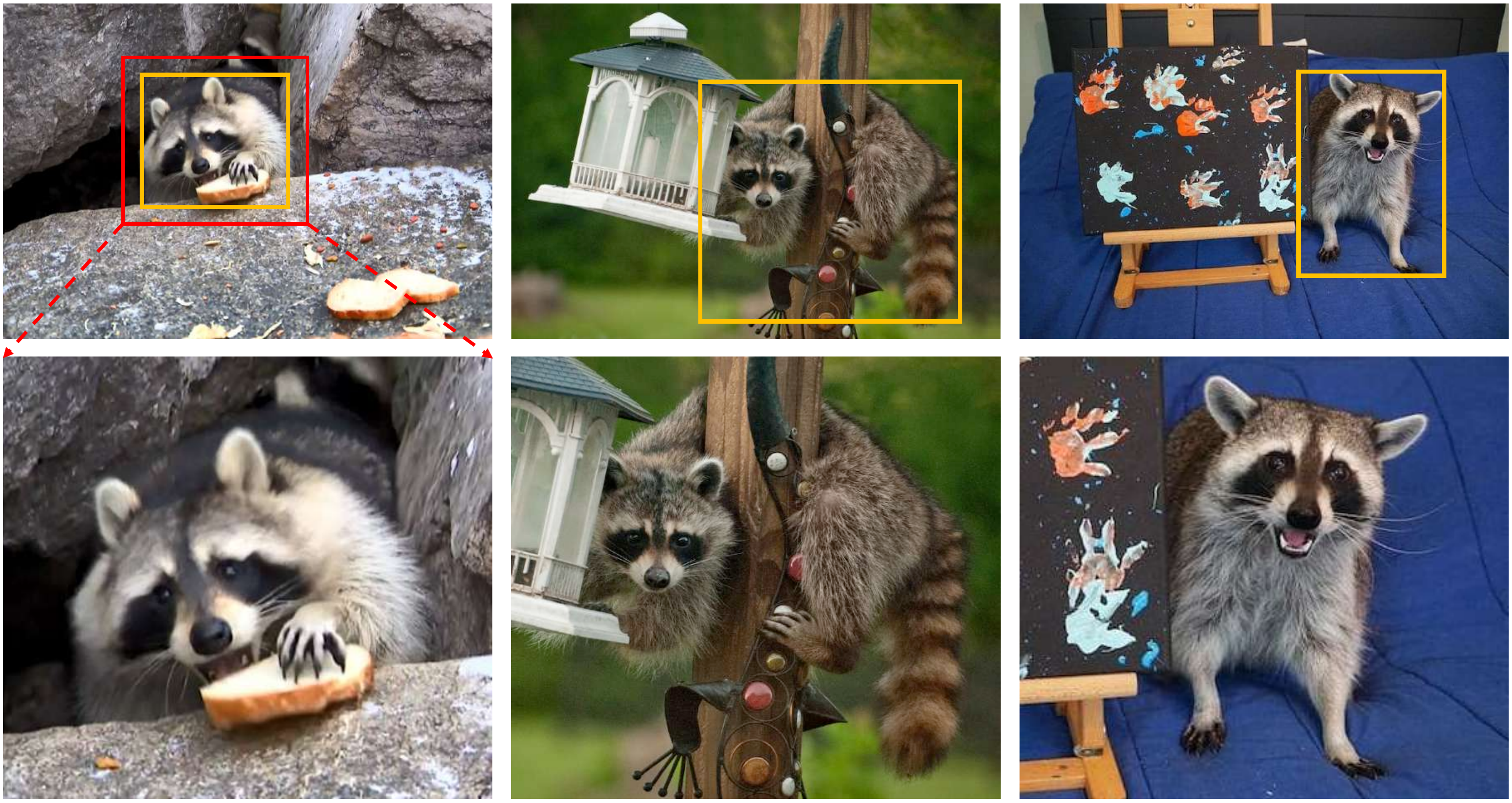}
\end{center}
\vspace{-10pt}
\caption{\textbf{Application 2.} Co-location based automatic thumbnails (``Raccoon'') generated by our \ourmodel.}
\label{fig:app2}
\end{figure}

\section{Downstream Applications}
Here, we show how the extracted co-saliency map can be utilized to generate high-quality segmentation masks for selected closely related downstream image processing tasks. 

\noindent{\textbf{Application \#1: Content-Aware Co-Segmentation.~}} 
Co-saliency maps have been previously used in pre-processing for unsupervised object segmentation. In our implementation, we first manually select a group of images from the internet by keyword search 
. Then, co-saliency maps are generated by our \ourmodel~to automatically mine the salient content of the specific group. Similar to Cheng \etal~\cite{cheng2014global}, we also utilize GrabCut~\cite{rother2012interactive} to obtain the final segmentation results. To initialize GrabCut, we simply choose adaptive threshold~\cite{peng2014rgbd} to binarize the saliency maps. 
Figure~\ref{fig:app1} shows the results of the content-aware object co-segmentation which should benefit  existing e-commerce applications requiring background replacement.

\noindent{\textbf{Application \#2: Automatic Thumbnails.~}}
The idea of paired-image thumbnails is derived from the seminal work~\cite{jacobs2010cosaliency}. With the same goal\footnote{Note that Jacobs \etal's work~\cite{jacobs2010cosaliency} is limited to the case of image pairs. 
}, 
we present a CNN-based photographic triage application which is valuable for sharing images with friends on the website. As shown in Figure~\ref{fig:app2}, we first generate the yellow box based on the co-saliency map obtained by our \ourmodel. Then, we simply enlarge the yellow box 
to get a larger red box. Finally, we adopt the collection-aware crops technique~\cite{jacobs2010cosaliency} to produce the results (2$^{nd}$ row). 

\section{Conclusion}

In this paper, we investigate a novel group collaborative learning framework (\ourmodel) for CoSOD. We find that group-level consensus can introduce effective semantic information to benefit the representation of both the \textit{intra-group compactness} and \textit{inter-group separability} 
for CoSOD.  Our experiments quantitatively and qualitatively demonstrate the advantage of our  {\ourmodel} which outperforms existing state-of-the-art models. In addition, our {\ourmodel}~achieves real-time speed (16ms) which can greatly benefit many applications such as co-segmentation, co-localization, and among others.

{\small
\bibliography{main}
}

\end{document}